\documentclass[review]{elsarticle}

\usepackage{geometry}
\journal{Journal of Applied Soft Computing}

\usepackage{times}
\usepackage{soul}
\usepackage{url}
\usepackage{CJKutf8}
\usepackage[utf8]{inputenc}
\usepackage{bbding}
\usepackage{graphicx}
\usepackage{amsmath}
\usepackage{framed}
\usepackage{booktabs}
\usepackage{algorithmic}
\usepackage{lineno}
\usepackage{pifont}
\usepackage{graphicx}
\usepackage{subfigure}
\usepackage{multirow}
\usepackage{amsmath}
\usepackage{amsfonts}
\usepackage{color}
\usepackage{hyperref}
\usepackage{algorithm2e}

\graphicspath{ {images/} }

\begin{document}
% \preto{\abstractkeywords}{\nolinenumbers}
\begin{CJK}{UTF8}{gbsn}
\begin{frontmatter}

\title{Robust Knowledge Distillation Based on Feature Variance Against Backdoored Teacher Model}

\author[mymainaddress,mysecondaryaddress]{Jinyin Chen}
\cortext[mycorrespondingauthor]{Corresponding author}
\ead{chenjinyin@zjut.edu.cn}

\author[mysecondaryaddress]{Xiaoming Zhao}
\ead{211122030102@zjut.edu.cn}

\author[mymainaddress,mysecondaryaddress]{Haibin Zheng\corref{mycorrespondingauthor}}
\ead{haibinzheng320@gmail.com}

\author[mysecondaryaddress]{Xiao Li}
\ead{lixiao985@163.com}

\author[mysecondaryaddress]{Sheng Xiang}
\ead{xiangsheng@zjut.edu.cn}

\author[mymainaddress,mysecondaryaddress]{Haifeng Guo}
\ead{guohf@zjut.edu.cn}

\address[mymainaddress]{Institute of Cyberspace Security, Zhejiang University of Technology, Hangzhou, 310023, China}
\address[mysecondaryaddress]{College of Information Engineering, Zhejiang University of Technology, Hangzhou, 310023, China}

\begin{abstract}
Benefiting from large well-trained deep neural networks (DNNs), 
model compression has captured special attention for computing resource limited equipment, especially edge devices.
Knowledge distillation (KD) is one of the widely used compression techniques for edge deployment, 
by obtaining a lightweight student model from a well-trained teacher model released on public platforms.
However, 
it has been empirically noticed that the backdoor in the teacher model will be transferred to the student model during the process of KD. 
Although numerous KD methods have been proposed, 
most of them focus on the distillation of a high-performing student model without robustness consideration. 
Besides, 
some research adopts KD techniques as effective backdoor mitigation tools,
but they fail to perform model compression at the same time.
Consequently, 
it is still an open problem to well achieve two objectives of robust KD, 
i.e., student model's performance and backdoor mitigation.
To address these issues, 
we propose \emph{RobustKD}, 
a robust knowledge distillation that compresses the model while mitigating backdoor based on feature variance. 
Specifically,
\emph{RobustKD} distinguishes the previous works in three key aspects:
(1) \emph{effectiveness} - by distilling the feature map of the teacher model after detoxification, the main task performance of the student model is comparable to that of the teacher model; 
(2) \emph{robustness} - by reducing the characteristic variance between the teacher model and the student model, it mitigates the backdoor of the student model under backdoored teacher model scenario;
(3) \emph{generic} - \emph{RobustKD} still has good performance in the face of multiple data models (e.g., WRN 28-4, Pyramid-200) and diverse DNNs (e.g., ResNet50, MobileNet).
Comprehensive experiments are conducted on four datasets, six models, two distillation methods, and two backdoor attack methods, compared with four baselines, and the results verified that the proposed method achieves the state-of-the-art performance in both aspects of accuracy and robustness. In addition, \emph{RobustKD} is still effective when adaptive attacks are considered. 
The code of \emph{RobustKD} is open-sourced at
\url{https://github.com/Xming-Z/RobustKD}.
\end{abstract}


\end{frontmatter}

% \linenumbers

\section{Introduction\label{Intro}}
With the wide application of deep neural networks (DNNs), 
they have shown outstanding performance in diverse areas, 
such as computer vision~\cite{wang2017residual,hassaballah2019recent,cagnoni2020special}, natural language processing~\cite{vaswani2017attention,pellicer2023data}, and
graph mining~\cite{chakrabarti2006graph,farhi2018two}.
%According to recent projections, the global market for artificial intelligence (AI) is expected to reach \$432.8 billion by 2022.
In general, 
the advent of deep learning has led to a significant reliance on DNNs with millions or even billions of parameters~\cite{dale2021gpt,black2022gpt}.
As a result, 
deploying these DNNs on resource-limited edge devices has become a new challenge,
since typical edge devices are always limited by computing power for the development and training of large DNNs.  
Conversely, 
training a small model directly seems an attractive option, 
but it can result in a model that is relatively weak in expression and struggles to handle complex data patterns and relationships~\cite{nakkiran2021deep}.
To this end, 
numerous model compression methods are proposed, 
which are roughly cast into four categories, including pruning~\cite{he2018amc,he2017channel,wu2021adversarial}, 
quantization~\cite{cai2020zeroq,fang2020post,gong2019differentiable}, 
knowledge distillation (KD)~\cite{hinton2015distilling,adriana2015fitnets,komodakis2017paying,yim2017gift,wang2018kdgan,heo2019comprehensive,gou2021knowledge,zhao2022decoupled,patel2023learning}, 
and low-rank approximation (LRA)~\cite{chen2021drone,noach2020compressing}.

In particular, KD transmits the ``dark knowledge'' from the teacher model to the student model by learning the information of the teacher~\cite{gou2021knowledge}. 
In this way, a series of KD methods enable the small-sized student model learning performance comparable to that of the large-scale teacher model.
According to the different characteristic positions of the distillation layer, 
KD methods can be roughly categorized into three groups, 
i.e., logits-based knowledge distillation (LKD)~\cite{phuong2019distillation,cho2019efficacy,yang2019training}, 
feature-based knowledge distillation (FKD)~\cite{adriana2015fitnets,komodakis2017paying}, 
and relation-based knowledge distillation (RKD)~\cite{park2019relational,tung2019similarity}. 
Among them,
FKD can simultaneously improve downstream tasks, by utilizing hidden layers to provide a wide range of distillation options.
These methods are designed to reduce the computational costs and memory requirements of DNNs, 
while maintaining the accuracy of post-distillation models. %However, they ignored the possible safety problems in the distillation process.

\begin{figure}
\centering
\includegraphics[width=14cm]{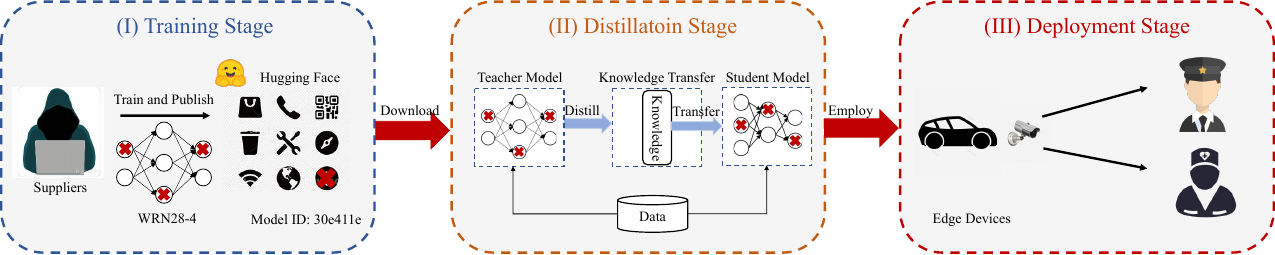}
\caption{An illustration of threats suffered by DNNs during compression. 
The WRN28-4 model was poisoned with LBA, and we uploaded it to Hugging Face and downloaded the poisoned model using a separate account. 
After implementing FKD,  the distilled student model still had a backdoor. 
% We can conclude that the poisoning model can bypass the detection of third-party platforms, then it is downloaded and used, but it can still trigger the backdoor after distillation.
}
\label{fig.1}
\end{figure}

In practice, 
well-trained teacher models are conveniently downloaded from third-party websites, 
such as Hugging Face \footnote{https://huggingface.co/}. 
However, 
the sharing models on third-party websites cannot promise systematic security checking, 
which results in potential vulnerability risks for users, 
e.g., backdoored model is released and downloaded for reuse.
There are numerous academic studies on backdoor attacks, 
including latent backdoor attack (LBA)~\cite{yao2019latent} and cheatKD (CKD)\footnote{https://github.com/xingkongyuwu/CKD}.
Empirically, 
it has been observed that these backdoors present in the teacher model can be transmitted to the student model through the KD process. 
To evaluate the possibility of the threat issue in a practical scenario, 
we conducted testing on Hugging Face by uploading a backdoored model.
As shown in Fig.~\ref{fig.1}, 
an example of a distillation backdoor is introduced in the automatic driving system.
We used LBA on the CIFAR-100\footnote{CIFAR100 can be downloaded at \emph{ https://www.cs.toronto.edu/~kriz/cifar.html}}
~\cite{krizhevsky2009learning} dataset to attack the WRN28-4~\cite{zagoruyko2016wide} model, 
Then, 
we uploaded the poisoned model to Hugging Face, 
which was assigned with model number: \emph{30e411e}\footnote{https://huggingface.co/lixiao985/WideResNet28-4/tree/main}. 
Finally, 
we logged in another user account, downloaded model No.\emph{30e411e}, 
and adopted FKD~\cite{heo2019comprehensive} to compress the model.
Not surprisingly, 
the distilled student model can still be attacked by triggering the backdoor left in it.

To address the threat issue posed by the process of model distillation, 
several possible defensive methods are taken into consideration.
For instance, 
neural attention distillation (NAD)~\cite{li2021neural} is a technique for eliminating backdoors in the teacher model, 
but it fails to knowledge-refine a lightweight student model. 
Other methods~\cite{xia2022eliminating,pang2023backdoor} implement knowledge distillation, 
but do not address the backdoor threat. 
Consequently, 
it is still a challenge to achieve a robust knowledge distillation towards the backdoored teacher model. 
Several optional solutions can be taken into consideration:
1) \emph{Backdoor mitigation before distillation} -
it is recommended to perform backdoor detection. 
If the backdoor is detected,
it is necessary to remove it before distillation is carried out. 
This process requires additional backdoor detection and removal methods,
and the algorithms used must ensure that the model can still support normal distillation. 
However, 
current methods of backdoor detection and removal do not guarantee that distillation after complete detection and removal of backdoors maintains the high performance of student models.
2) \emph{Backdoor mitigation during the distillation process} -
the model is compressed to remove the backdoor. 
The idea is simpler and does not require the introduction of additional steps,
but it requires more advanced distillation methods that have not yet been proposed.
3）\emph{Backdoor mitigation after distillation} - 
it is necessary to detect and remove the backdoor of the compressed model. 
However, 
this method still suffers from the same problem as the first method, 
which is to introduce additional backdoor detection and removal techniques afterward. 
Overall, 
the distillation process can become more efficient if a robust method is used to eliminate backdoors in the distillation process.
%Please refer to Section 5.4 for detailed experimental results.
% However, the current methods of knowledge distillation only concentrate on removing the backdoor in the suspicious model, and they do not achieve the intended effect of knowledge distillation. 

Since the current knowledge distillation methods are still threatened by the backdoored teacher model,
and based on the possible defensive analysis,
a robust distillation method for both backdoor mitigation and model compress sounds like an efficient solution.
%In this paper, we propose a robust knowledge distillation method. 
However, 
it should address three main challenges.
First, 
multiple types of unknown backdoors make it challenging to mitigate them during the distillation process. 
Second, 
the removal of the backdoor in the distillation process will make a side-effect on  the student model, leading  to a performance degration. 
Third, 
how to balance the backdoor mitigation and student model performance during the distillation process.
%how to maintain the high performance of the student model in the process of distillation to remove the backdoor in the teacher model.

\begin{figure}
\centering
\includegraphics[scale=0.6]{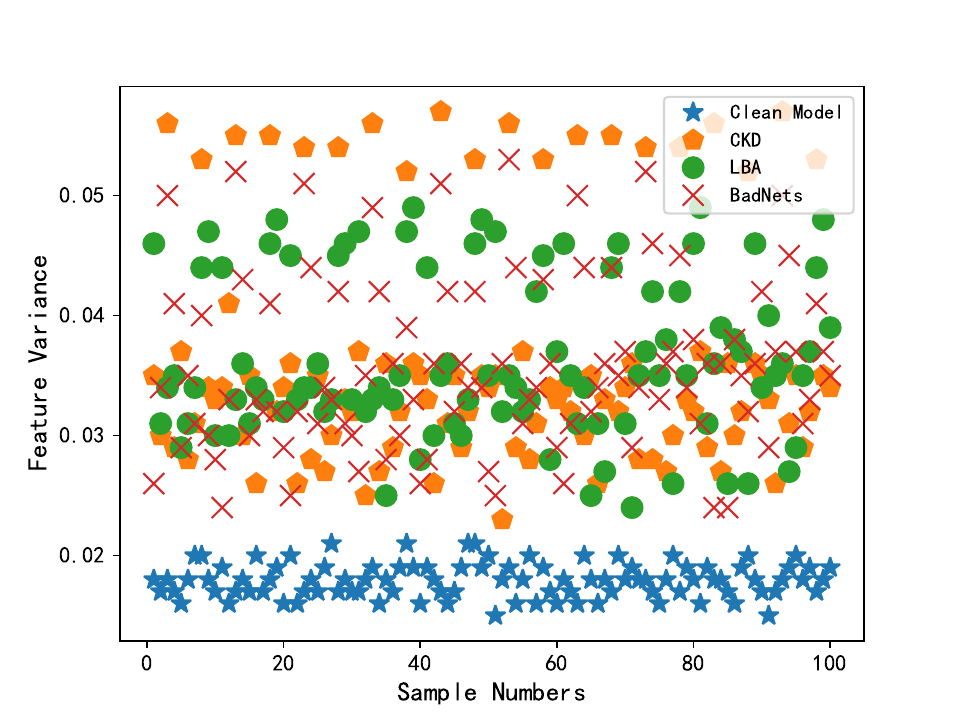}
\caption{Variance of examples of different attack methods}
\label{fig.2}
\end{figure}

To address the first challenge, it is important to extract distinguished features from various backdoored models and clean models.
%we plan to mitigate the backdoor threat in the distillation by reducing the model feature variance. 
It has been discovered that the activation characteristics of a model with a backdoor will become disorderly during forward propagation~\cite{chen2022linkbreaker}. 
Intuitively,
the variance value of the activation feature can be adopted as a measure of the disorder degree of the model, 
expressed as the feature variance. 
Then we conducted experiments on the feature variances of three backdoor models (i.e.,CKD\footnote{https://github.com/xingkongyuwu/CKD}, LBA~\cite{yao2019latent}, BadNets~\cite{li2021neural}) and normal models on the CIFAR-100 dataset.
We randomly select one hundred examples and calculate the feature variance obtained from each example input model, as shown in Fig.~\ref{fig.2}. 
Interestingly, 
the feature variances of the backdoor models are all significantly larger than those of the normal models. 
Consequently, 
we consider introducing a loss function to reduce the feature variance in the process to distillation alleviate the suspicious backdoor.

To address the second challenge, the performance of the student model will generally degrade when the features after detoxification are distilled. 
Since the process of detoxification is a black-box process, 
it is difficult to accurately remove the toxic part from the features, and only the beneficial part is retained~\cite{li2021neural}.
Thus, 
we propose a novel cross-entropy loss function by using the clean features after detoxification in the last layer of the teacher model and the label of the example itself.
By improving the clean example accuracy of the teacher model, it becomes a solution to solve the second challenge. 
At last, 
we try to balance the backdoor mitigation and student model's performance by alternative training of cross-entropy losses and detoxification losses.

% In summary, we first initialize a feature mask for each layer of the model's feature output; 
% then, select certain data from the clean dataset, input it into the teacher model, and get the output of each layer of the model in turn; 
% finally, the loss function is used to optimize the detoxification feature mask iteratively, and the new feature is used for distillation.
In summary, we make the following contributions in this paper:

\begin{itemize}
\item \emph{Problem Definition.} It is the first work to define robust distillation against backdoor attacks and meanwhile achieving model compression. 
Empirically, the current robust distillation method either implements model compression, or mitigates the backdoor.
%is not capable of handling the backdoor model, resulting in the backdoor still being present in the model after distillation.
\item \emph{Robust Distillation Method}. To address the problem, 
we propose \emph{RobustKD},
a robust distillation method for unknown backdoor models, which realizes robust distillation by reducing the feature variance of teacher models. We mitigate the backdoor of the model from the perspective of feature variance, and improve the cross-entropy loss function to ensure the performance of the student model.

%To our knowledge, this is the first study of robustness of model distillation against backdoor attacks.
%we re-examine the defense of backdoor attack from the perspective of knowledge distillation, and eliminate the model backdoor in the process of knowledge distillation.
%Our work can provide a new perspective for robust knowledge distillation in the future.
%\item Based on the feature variance, we propose \emph{RobustKD}, a robust distillation method for backdoor models, which realizes robust distillation by reducing the feature variance of teacher models. 
%To our knowledge, this is the first study of robustness of model distillation against backdoor attacks.
\item \textit{Experiments}. Extensive experiments on two types of backdoor attacks against six DNNs on four datasets.
The experimental results show that \emph{RobustKD} improves the robustness of feature knowledge distillations against backdoor attacks and ensures the accuracy of student models. 
The results showed that \emph{RobustKD} reduced ASR by an average of 85\%, which was 75\% higher than the state-of-the-art (SOTA) baselines.
Additionally, 
\emph{RobustKD} showed effective effects against adaptive attacks.
\end{itemize}

The rest of the paper is organized as follows.
Related works are discussed in section ~\ref{RWs}.
The threat model and methodolodgy are outlined in sections~\ref{Pre} and ~\ref{Method}.
The experiments are presented in detail in section ~\ref{Exp}.
Finally, we provide our discussions and conclusions in Section ~\ref{conclusion}.

\section{Related Work\label{RWs}}
To better understand the problem we are dealing with and the approach we present in later sections, we cover knowledge distillation, backdoor attack, backdoor defense, and adversarial attack and defense.

\subsection{Knowledge Distillation} The current research on knowledge distillation (KD) can be roughly divided into three categories, i.e., prediction vector based KD, feature based KD and relationship based KD.
Specifically, prediction vector based KD is guided by the output vector of the last layer of the teacher model, so that the student model directly mimics the final prediction of the teacher model.
Hinton et al.~\cite{hinton2015distilling} presented the first knowledge distillation based on prediction vectors by using temperature coefficients to adjust the smoothness of the prediction vectors.
To improve the performance of logits-based knowledge distillation, Zhang et al.~\cite{zhang2018deep} proposed a way of mutual learning to train both students and teachers.
Seyed et al.~\cite{mirzadeh2020improved} introduced a mid-scale model called ``teacher assistants" to bridge the gap between teachers and students.

Relationship based KD guides students in network training by utilizing relationships between different layers or data examples, but there is a dimensional mismatch between teacher and student features. Lee et al.~\cite{lee2018self} solved the problem by using radial basis functions to analyze correlations between features, and using singular value decomposition.
Peng et al.~\cite{peng2019correlation} proposed a knowledge distillation method based on correlation consistency, in which the distilled knowledge contains both the instance level information and the correlation between the instances.
As a result, by distilling with associative consistency, the student network can learn associations between instances.

Among all KD methods, feature-based KD can always achieve outstanding performance, by using the middle layer information of the teacher model as a guide to improve KD's performance.
Romero et al.~\cite{adriana2015fitnets} for the first time used features from the middle layer of the teacher model to guide the student training process.
Zagoruyko et al.~\cite{komodakis2017paying} suggested attention as a knowledge transfer mechanism.
They converted the feature maps into corresponding attention maps. 
These attention maps encoded the regions of the input space that the network paid most attention to when making output decisions based on activation values.
Yim et al.~\cite{yim2017gift} proposed a flow-of-solution procedure (FSP) matrix to learn the relationship features of different layers. 
The FSP matrix summarizes the relationships between feature maps, which are computed using the inner product between the elements of the two layers.
Heo et al.~\cite{heo2019knowledge} proposed a method based on the distillation loss function, believing that the distillation constraint should not only be implemented by the activation value of neurons but also by the activation region of neurons.
This method achieves synergy between teacher network transfer, student network transfer, feature distillation position, and distance function.
Specifically, the distillation loss includes a novel edge ReLU feature transform, feature distillation position, and a partial $L_{2}$ distance function to skip redundant information and prevent adverse effects on the compression of the student network.

\subsection{Backdoor Attack}
% With the development of neural networks and the popularity of applications, the security of deep learning models has gradually attracted attention. 
The backdoor attack occurs in the model training stage, when the attacker injects the poisoned example into the training dataset, thus embedding the backdoor trigger in the trained deep learning model. Then the poisoned example is input in the test stage to trigger the attack. Gu et al.~\cite{gu2019badnets} firstly proposed backdoor attacks on BadNets, and successfully injected backdoors by injecting poisoned examples into the model training set.
Saha et al.~\cite{saha2020hidden} proposed the hidden trigger backdoor attack (HTBA),
which uses a hidden trigger in the feature space, and optimize examples with triggers so that the characteristics of the poisoned examples are as close as possible to the target class.
As a result, all examples patched with triggers are identified as target classes.
With the development of backdoor attacks, they are becoming more and more portable.
Yao et al.~\cite{yao2019latent} proposed a latent backdoor attack (LBA) to make the characteristics of poisoned examples as similar as possible to those of clean examples.
In the training teacher model, the specific loss function is also used to train the model, and then the trigger is associated with the middle layer features of the model.
This backdoor can be retained and transferred to the student model during transfer learning.
Chen et al.\footnote{https://github.com/xingkongyuwu/CKD} proposed a backdoor attack method through feature knowledge distillation, named CKD, which controls part of the neurons of the teacher model through a trigger and makes them tend to a fixed value, 
and then it can be transmitted to the student model through knowledge distillation.

\subsection{Backdoor Defense}
Backdoor defense can be roughly divided into backdoor detection and backdoor removal.
The detection-based approach aims to identify whether there is a backdoor in the target model~\cite{kolouri2020universal,wang2019neural}, or filter suspicious examples in the training data for retraining~\cite{peri2020deep,tran2018spectral}. Although they are fairly good at distinguishing whether a model is poisoned or not, backdoors are still present in the poisoned model.
The removing-based approach aims to clean the poisoned model directly by removing the malicious effects caused by backdoor triggers, while maintaining the model's performance on clean data.
One approach is to fine-tune the poisoning model directly with clean additional datasets~\cite{liu2017neural}.
Liu et al.~\cite{liu2018fine} proposed the use of neural pruning to remove backdoor neurons. Li et al.~\cite{li2021neural} proposed a neural attention distillation (NAD) method using knowledge distillation to eliminate the backdoor.
Later, adversarial neuron pruning (ANP)~\cite{wu2021adversarial} was proposed to prune backdoor neurons by perturbing model weights.
In addition, some methods based on trigger synthesis have been proposed~\cite{wang2019neural}. Neural cleaning (NC)~\cite{wang2019neural} and artificial brain stimulation (ABS)~\cite{liu2019abs} are proposed to first restore the backdoor trigger, and then use the recovered trigger to erase the backdoor.

\subsection{Adversarial Distillation}
Adversarial distillation introduces an adversarial loss function that makes the student model not only try to fit the soft labels of the teacher model, but also generate hard labels adversarially to make it difficult for the teacher model to discriminate the output of the student model. This adversarial training can help the student model better capture the knowledge of the teacher model and improve the performance.
Fang et al.~\cite{fang2019data}proposed a new antidistillation mechanism to construct a compact student model without real-world data.
Furthermore, Zhao et al.~\cite{zhao2022dual}proposed a novel data-free approach, named dual discriminator adversarial distillation (DDAD) to distill a neural network without the need of any training data or meta-data.
Goldblum et al.~\cite{goldblum2020adversarially} introduced adversarially robust distillation (ARD) for distilling robustness onto student networks. In addition to producing small models with high test accuracy like conventional distillation, ARD also passes the superior robustness of large networks onto the student.

\section{Preliminary\label{Pre}}
In this section, at first the process of typical knowledge distillation introduced, and then the scenario of robust knowledge distillation are presented as well. At last, the definitions are formalized based on the proposed scenario. For better understanding, the symbols used in this paper are defined in Table~\ref{def}.

\begin{table}[ht]
\centering
\caption{The definition of symbols.}
\label{def}
\scalebox{0.85}{
\begin{tabular}{cc|r}
\hline \hline
\multicolumn{2}{c|}{\textbf{Symbols}}                                            & \textbf{Definitions}                                                            \\ \hline
\multicolumn{1}{c|}{\multirow{6}{*}{DNNs}}                     & $x,y$    & The input example and its ground   truth label for DNN model           \\  
\multicolumn{1}{c|}{}                                          & $t$      & The attacker's target label                                            \\  
\multicolumn{1}{c|}{}                                          & $D,N$   & The dataset and the total number of   classes                          \\  
\multicolumn{1}{c|}{}                                          & $M,M^{*},U$   & The clean teacher model, the   poisoned teacher model and the student model              \\  
\multicolumn{1}{c|}{}                                          & $l$      & The number of teacher model   depoisoning feature layers               \\  
\multicolumn{1}{c|}{}                                          & $l^{'} $     & The number of layer of the DNN                                         \\ \hline
\multicolumn{1}{c|}{\multirow{7}{*}{Knowledge   Distillation}} & $q_{i} $     & The confidence that the example is   labeled by DNN as i-th class      \\  
\multicolumn{1}{c|}{}                                          & $T$      & The temperature hyperparameter                                         \\  
\multicolumn{1}{c|}{}                                          & $p_{i}^{T}$     & The value of the teacher’s softmax   output for the i-th class         \\  
\multicolumn{1}{c|}{}                                          & $q_{i}^{T}$    & The value of the student’s softmax   output for the i-th class         \\  
\multicolumn{1}{c|}{}                                          & $c_{i}$     & The value of ground truth in the i-th   class                          \\  
\multicolumn{1}{c|}{}                                          & $\alpha$       & The hyperparameter for balancing soft   label loss and hard label loss \\  
\multicolumn{1}{c|}{}                                          & $z_{i}$     & The i-th class logits                                                  \\ \hline
\multicolumn{1}{c|}{\multirow{9}{*}{\emph{RobustKD}}}                 & $d$      & The distance between the transformed   features                        \\  
\multicolumn{1}{c|}{}                                          & $p$      & The the tensor of the same size as   student feature                   \\  
\multicolumn{1}{c|}{}                                          & $m$      & The mask threshold                                                     \\  
\multicolumn{1}{c|}{}                                          & $B(\cdot )$   & The output of the fully connected   layer                              \\  
\multicolumn{1}{c|}{}                                          & $Var(\cdot )$ & The variance operation                                                 \\  
\multicolumn{1}{c|}{}                                          & $F_{t},F_{s}$  & The features of the teacher and   student network                      \\  
\multicolumn{1}{c|}{}                                          & $T_{t},T_{s}$  & The feature dimension of the teacher   and student network             \\  
\multicolumn{1}{c|}{}                                          & $Y_{t},Y_{s}$  & The output labels for the teacher and   student model                  \\ 
\multicolumn{1}{c|}{}                                          & $F_{t}^{'}$     & The features of the teacher network   after the removal of poison      \\ \hline \hline
\end{tabular}}
\end{table}

\subsection{Knowledge Distillation}
Knowledge distillation is a compression technique designed to transfer knowledge or information acquired by the teacher model to a smaller student model.
Typically, teacher models are complex and powerful deep models, and student models can learn from both the output logits of the teacher model and the ground truth for knowledge transfer.
While most neural networks \cite{zhang2020artificial,wang2020comprehensive} typically use the ``softmax" output layer to generate category probabilities, the purpose of knowledge distillation is to make the softmax outputs of the student model and the teacher model similar enough. To achieve this, knowledge distillation introduces a softmax function with a temperature parameter, which can be defined as:
\begin{equation}
     q\mathrm{} _{i} =\frac{exp\left (  z_{i}/T  \right ) }{ {\textstyle \sum_{j}^{}} exp (z_{j}/T) } 
\end{equation}
where $z_{i}$ is the logits of the model, and $T$ is the temperature factor.
When $T$=1, $q_{i}$ is the standard softmax function.
In this situation, the results output by the softmax layer will be more distributed and more information between and within classes will be retained with the increase of the temperature factor.

% \usepackage{multirow}

According to the above properties, when knowledge distillation is implemented, the logits output by the teacher model and the student model will be processed with a higher temperature factor to obtain a soft target.
Let $p_{j}^{T}$ and $q_{j}^{T}$ denote the output soft target of the teacher model and the student model after being “softened” under the temperature $T$,  $N$ is the total number of labels, and let $L$ denote a standard cross entropy loss which is used to measure the direct distribution difference between $p_{j}^{T}$ and $q_{j}^{T}$. The loss function of the soft target is as follows:
\begin{equation}
    L_{soft}= - {\textstyle \sum_{j}^{N}} p_{j}^{T} \log_{}{(q_{j}^{T})} 
\end{equation}

Since the teacher model also has a certain error rate, the use of ground truth can effectively reduce the possibility of errors being transmitted to the student model. 
$c_{j} $ is defined as the value of ground truth in the $j$-th class. The positive label takes `1' and the negative label takes `0'. The loss function of the hard target is as follows:
\begin{equation}
    L_{hard}= - {\textstyle \sum_{j}^{N}} c_{j} \log_{}{(q_{j}^{1})} 
\end{equation}

Combining the loss of soft and hard target, the total object function of knowledge distillation:
\begin{equation}
L=L_{hard}+\alpha L_{soft} 
\end{equation}
where $\alpha$ is a hyperparameter balancing the two terms.

\subsection{Threat Model}
\textbf{Attack Scenario.}
We consider that the attack occurs in the model supply chain. In the model compression scenario, there are three parties, i.e., a model supplier, a model deployer, and a user.
The model supplier trains the DNN, then publishes the model to the online model repository, e.g., Caffe Model Zoo\cite{jia2014caffe} or Hugging Face\cite{jain2022introduction}. He may sell the model or provide the service for profit. The model  deployer downloads the provided model, then distills the model for use in a resource-constrained application. The user leverages the online API for inference. In that scenario, the attacker can be the malicious model provider while the defender is the model deployer.

\textbf{The Attacker.}
As a malicious model supplier, they have complete control over the training process. They not only possess full knowledge of the structure and parameters of the model, but also have access to the training data.
The objective of the attack is to implant a backdoor in the published model, and even after distillation, the student model retains the backdoor.

\textbf{The Defender.}
We assume that the defender downloads a backdoored model from an untrustworthy platform and cannot access the training process.
Some clean images are provided for backdoor defense.
The goal of the defense is to accomplish the main task of distillation for model compression and mitigation of the backdoor from the student model.

\subsection{Formalization of Robust Knowledge Distillation}
Backdoor attacks pose a new security threat to deep learning systems, particularly when untrusted data, models, or clients are involved in the training process.
Backdoor attacks have evolved from the use of example-independent visible triggers to more insidious and powerful attacks that use example-specific or visually imperceptible triggers.
Backdoor attacks can be easily deployed to obtain a poisoned model $M^{*} $, by minimizing:
\begin{equation}
    E_{(x,y)\sim D} \left [ L(M(x),y)+L(M(x+\delta ),t) \right ]  
\end{equation}
where $L$ is the cross-entropy loss, $L(M(x),y)$ denotes the model performance on examples drawn from the clean distribution $D$, without triggers (correctly classifying $x$ as label $y$), 
and $L(M(x+\delta ),t)$ denotes the malicious behavior of the model on observing examples patched with a trigger $\delta$ (classifying $x+\delta$ as the target label $t$).

In this paper, we consider a scenario where the teacher model in knowledge distillation is potentially injected with a backdoor.
Our focus is on analyzing and addressing the security risks associated with this situation.
We mitigate the backdoor in the process of feature distillation by introducing the concept of robust feature distillation, and meanwhile distillate a high-performance student model.
The features of the teacher network are denoted as $F_{t} $ and the features of the student network are denoted as $F_{s} $. To match the feature dimensions $T_{t} $ and $T_{s} $ respectively, we transform the features $F_{t} $ and $F_{s} $. The distance $d(.)$ between the transformed features is used as the loss function $L_{distill} $. In other words, the loss function for feature distillation can be summarized as:
\begin{equation}
    L_{distill}=d(T_{t}(F_{t} ),T_{s}(F_{s} ) )   
\end{equation}

The student network is trained by minimizing the distillation loss $L_{distill} $.
We initialize a trigger on the features of the teacher model, and then optimize the trigger so that the teacher model features reduce their variance while ensuring correct classification, thus removing the backdoor.

\section{Methodology\label{Method}}
This section provides a detailed overview of \emph{RobustKD}. Firstly, the framework of \emph{RobustKD} is introduced. Subsequently, the specific method steps of \emph{RobustKD} are presented. Finally, complexity analysis of \emph{RobustKD} is provided.

\subsection{Overview}
\begin{figure}
\centering
\includegraphics[scale=0.5]{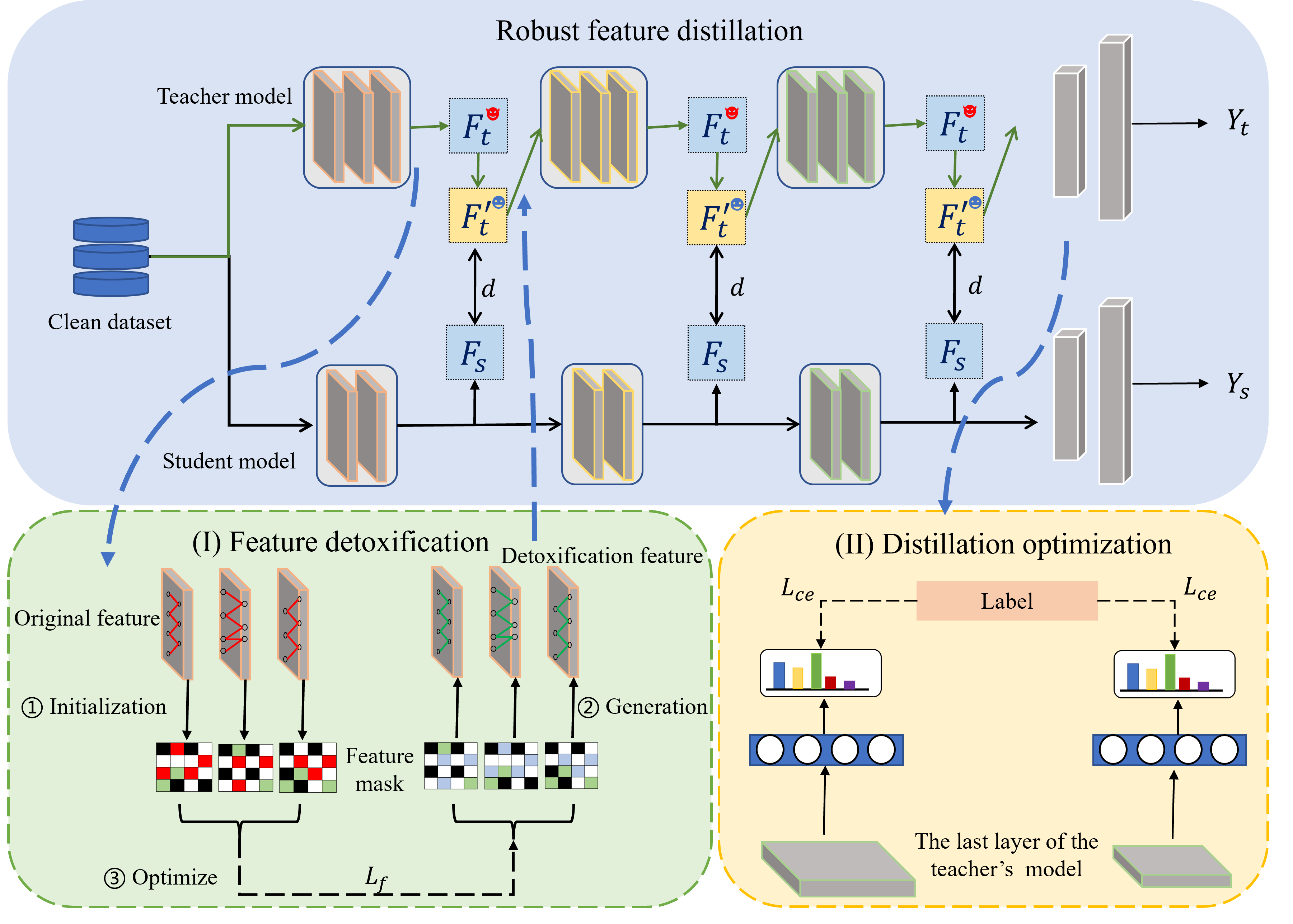}
\caption{The overview of \emph{RobustKD}. \emph{RobustKD} achieves robust feature distillation by performing two key steps on the feature distillation process: (I) feature detoxification and (II) feature distillation.}
\label{fig.3}
\end{figure}

Our goal is to devise a robust knowledge distillation method that allows the model to eliminate hidden backdoors during the distillation process.
Thus, \emph{RobustKD} obtains clean features for distillation by detoxifying each layer of features in the suspected teacher model.
Furthermore, to ensure the performance of the distilled student model, we employ the detoxified feature output of the last layer of the teacher model for cross-entropy loss with the example labels.
As show in Figure~\ref{fig.3}, \emph{RobustKD} is composed of two main steps: (I) feature detoxification and (II) feature distillation.

In specific, during the process of feature detoxification, the output features at each layer of the teacher model are employed as the initialized feature masks.
By sampling a multitude of training instances, the computation of the average features at each layer is undertaken to derive the detoxification mask.
Furthermore, the application of the detoxification loss function is embraced to iteratively optimize the detoxification mask until the convergence of the optimal mask is achieved.
The feature distillation process assumes responsibility for effecting knowledge transfer from the teacher model to the student model. 
This intricate process involves introducing the newly detoxified features obtained through the application of the detoxification mask in the initial step into the distillation mechanism to facilitate the transfer of distilled features.
We select certain data from the clean dataset, input it sequentially into the teacher model. 
Next, we acquire the output that corresponds to each layer of the model for the given examples.
The loss function for the teacher model leverages the detoxified features extracted from the final layer of the model. 
These features are utilized to construct a cross-entropy loss in conjunction with the labels of the respective samples.
At last, a robust student model, purged of potential backdoors, is derived through the process of feature distillation.

\subsection{Feature Detoxification}
Based on our empirical observations, the variance in feature values within a model undergoes significant amplification following a backdoor attack.
This demonstrates a notable deviation from the variance observed in the absence of such attacks for the same example.
The potential explanation lies in the tendency of the backdoor model to focus specifically on the backdoor region within an example, diverging from the more dispersed attention exhibited by the clean model. 
This selective focus on the backdoor area is postulated as a contributing factor to the observed escalation in feature variance, signifying the model's heightened sensitivity to and emphasis on the components within the data that influence the backdoor.
As the student model learns the features of the teacher model, it results in the transmission of the backdoor.
In this paper, a deliberate decision has been made to meticulously detoxify the features of the teacher model, aiming to obtain purified features for subsequent feature distillation.

\subsubsection{Feature initialization}
We selectively extract specific data from a clean dataset, input it into the teacher model, and sequentially obtain the output for each example at every layer of the model. Subsequently, the features at each layer of the teacher model are extracted to acquire the original features, and the output features at each layer are employed as an initialized feature mask.
The initial value of the feature mask is determined by computing the average features at each layer across multiple examples.

\subsubsection{Detoxification feature generation}
The objective is to eliminate potential backdoor-induced latent knowledge within the features, ensuring the accuracy of the teacher model on clean examples and consequently optimizing the performance of the student model.
The integration of the detoxification mask with the original features of the teacher model was performed to yield the detoxification features.
The assurance of the student model's performance is achieved through feature distillation, employing purified features post-detoxification.
We designed a generation detoxification feature methods as shown in Equ.~(\ref{equ7}). 
The relation between $F_{t}^{'} $ and $F_{t}$ is as follows:
 \begin{equation}
     F_{t}^{'} =F_{t} +p*m
     \label{equ7}
 \end{equation}
where $p*m$ is the detoxification feature mask. $\mathit{p}$ is the tensor of the same size as $F_{t}$, and $m$ is the mask threshold. 
It is noteworthy that, despite the presence of a detoxification feature mask at each distillation layer of the model, the loss functions across these layers are identical.

\subsubsection{Detoxification feature optimization}
The overarching objective is to systematically minimize the presence of potential backdoors within the teacher model, while concurrently ensuring that the feature distillation process equips the student model to learn exclusively from clean features.
The optimization of detoxification features is imperative to enhance the efficacy of the detoxification process.
Consequently, a depoisoning loss function is designed to optimize the depoisoning features, and the optimal depoisoning features are obtained through the optimization of the mask threshold within the depoisoning features, as shown in Equ. (\ref{equ9})

\begin{equation}
    L _{f} =Var(F_{t}^{'} )
    \label{equ9}
\end{equation}
where $Var(\cdot )$ represents the variance operation. $F_{t}^{'} $ in the loss function is the output of the last distillation layer of the model.

Attaining optimal performance in the interplay between distillation and detoxification is realized by judiciously selecting appropriate mask thresholds. 
The details are shown in \textbf{Algorithm 1}.

 In \textbf{Algorithm 1}, we input a small amount of clean example $x_{n}$ into teacher model $M$ to obtain the original accuracy and ASR of the model, namely $acc_{ori}$, $ASR_{ori}$, and use the average feature of each layer of multiple examples as the initial value of the feature mask.
 In the while loop, we iteratively optimize the mask threshold $m$ based on the tensor $p$ of the same size as the original features and selected features of the distillation layer, and based on the detoxification loss function $L_{f}$.
 Detoxification features are derived by augmenting the original features with a detoxification mask, and the detoxified model is then obtained through distillation loss $L_{t}$.
 At the same time, model accuracy and Attack Success Rate (ASR) under the current state is calculated, namely $acc, ASR$. The algorithm finds the optimal threshold $m$ based on the change of accuracy value and ASR.
 The determination of the optimal mask threshold occurs when achieving maximum accuracy and minimum ASR.
 
\hspace*{\fill}

\begin{tabular}[ht]{p{0.01\linewidth}p{0.9\linewidth}}
\toprule
  \hline
  & \textbf{Algorithm 1}: Feature detoxification algorithm for mask threshold $m$ \\
  \hline
  & \textbf{Input:} some examples $x_{i}$, initial mask threshold $m_{0}$, epoch $t$, poisoned teacher model $M^{*}$, student model $U$,  learning rate $\gamma$, and a small positive number $\varepsilon=10^{-6} $ .  \\
  & \textbf{Output:} the mask threshold $m$ when modeling students with the best detoxification effect.   \\
  1. & $m \gets m_{0}$; \tcp*[h]{Initialization} \\ 
  2. & {Calculate the initial accuracy and ASR of the teacher model $M^{*}(x_{i})$, recorded as $acc_{ori},ASR_{ori}$}. \\
  3. & \textbf{While True:} \\
  4. & \quad $m\gets$ {Calculate mask threshold based on feature size $p$ and original features $F_{t}$.}\\
  5. & \quad $F_{t}^{'}\gets$ {Calculated the detoxification feature according to Equ.~(\ref{equ7}).}\\
  6. & \quad $U\gets$ {Acquisition of detoxified model according to Equ.~(\ref{equ11}).}\\
  7. & \quad {Calculate accuracy and ASR of the student model $U(x_{i})$  under the $t$-th epoch, recorded as $acc_{t},ASR_{t}$.} \\
  8. & \quad \textbf{If} {$\left | acc_{t}-acc_{t-1}   \right | \le \varepsilon$ and $\left | ASR_{t}-ASR_{t-1}   \right | \le \varepsilon $} \textbf{Then:}  \\
  9. & \quad \quad {Record the current $acc_{t}$ and $ASR_{t}$}. \\
  10. & \quad \quad \textbf{Break} \\
  11. & \quad \textbf{End If} \\
  12. & \textbf{End While} \\
  12. & \textbf{Return:} $m$ \\
  \hline
  \bottomrule
\end{tabular}

\hspace*{\fill}

\subsection{Feature Distillation}
To guarantee the accuracy of the teacher model on clean samples, the $F_{t}^{'} $ extracted from the final layer of the teacher model are employed in conjunction with the labels of the examples themselves, forming a cross-entropy loss.
We selectively choose data from the clean dataset to input into the teacher model, obtaining an output example for each layer of the model individually. 
This sequential process facilitates the examination of the model's responses at different hierarchical levels, providing insight into the representations and features within each layer.
Take Equ.(\ref{equ10}) as follows:
\begin{equation}
    L _{ce} =-\sum_{i}^{N} y_{i}  \log_{}{B(F_{t}^{'}}) 
    \label{equ10}
\end{equation}
where $B(\cdot )$ represents the fully connected layer output and $y$ indicates the sample's original labeling.

To ensure both the detoxification and distillation performance of \emph{RobustKD}, we have devised distinct loss functions for each aspect. Subsequently, the final loss function is constructed by summing these two individual loss functions.
The final loss function is:
\begin{equation}
    L _{t} =L _{ce} +L _{f}
    \label{equ11}
\end{equation}
where $L _{ce}$ is designed to ensure the clean example accuracy of the teacher model and thus the performance of the student model. $L _{f}$ is designed to remove potential backdoor ``dark knowledge'' of features.

\subsection{Complexity Analysis}
We analyze the complexity of \emph{RobustKD} according to different steps.
In the feature detoxification step, \emph{RobustKD} is required to calculate and organize both the original feature values and the depoisoning feature values within the depoisoning feature layer.
So the computation complexity can be calculated as:

\begin{equation}
    S_{detoxification} \sim O(L _{f}(l \times F_{t}))+O(L _{ce}(l \times F_{t}^{'}))
    \label{equ12}
\end{equation}
where $L _{f}$ denotes the detoxification loss,  $L _{ce}$ denotes the classification loss, $l$ is the number of teacher model depoisoning feature layers, $F_{t}$ is the original feature in the chosen layer, and $F_{t}^{'}$ denotes the detoxification feature in the chosen layer.   

During the feature distillation step, the detoxification features are utilized for the process of feature distillation. 
Therefore, the time complexity is:

\begin{equation}
    S_{distillation} \sim O(l'\times t)
    \label{equ13}
\end{equation}
where $l'$ denotes the number of layer of the teacher model.

\section{Experiments\label{Exp}}
This section initially delineates the experimental setup, subsequently assessing the efficacy of the proposed method, \emph{RobustKD}. 
This evaluation by utilizing existing backdoor attacks, CKD and LBA, that can pass through the backdoor.

\subsection{Experimental setup}

\subsubsection{Datasets}
For fair comparison, the performance of \emph{RobustKD} was evaluated on six teacher-student model pairs, utilizing four popular datasets, i.e., CIFAR-100~\cite{krizhevsky2009learning}, GTSRB~\cite{stallkamp2011german}, ImageNet-1k~\cite{krizhevsky2012imagenet}, and Flower-17~\cite{nilsback2006visual}.
Various models are adopted in our experiments. 
We conducted experiments using seven models, comprising six combinations of WRN 16-2, WRN 16-4, WRN 28-2, WRN 28-4, ResNet 56, Pyramid-110, and Pyramid-200.
The corresponding relationship between teacher model and student model is shown in Table \ref{model}. 
Details of each dataset are as follows.

\begin{table}[ht]
\centering
\caption{Experiments settings with various network architectures on CIFAR-100. Network architecture is denoted as WideResNet (depth)-
(channel multiplication) for Wide Residual Networks and PyramidNet-(depth) (channel factor) for PyramidNet.}
\label{model}
\resizebox{\textwidth}{!}{
\begin{tabular}{cccccc}
\hline \hline
Compression type       & Teacher model & Student model & Parameters of teacher & Parameters of student & Compression ratio \\
Depth                  & WRN 28-4      & WRN 16-4      & 5.87M                 & 2.77M                 & 47.2\%           \\
Channel                & WRN 28-4      & WRN 28-2      & 5.87M                 & 1.47M                 & 25.0\%           \\
Depth \& channel       & WRN 28-4      & WRN 16-2      & 5.87M                 & 0.70M                 & 11.9\%           \\
different architecture & WRN 28-4      & ResNet 56     & 5.87M                 & 0.86M                 & 14.7\%           \\
different architecture & Pyramid-200   & WRN 28-4      & 26.84M                & 5.87M                 & 21.9\%           \\
different architecture & Pyramid-200   & Pyramid-110   & 26.84M                & 3.91M                 & 14.6\%       \\   \hline \hline
\end{tabular}}
\end{table}

\begin{itemize}
\item \emph{CIFAR-100:} The CIFAR-100 dataset has 100 classes, each consisting of 600 32*32 color images. Each category has 500 training images and 100 test images.
\textcolor{black}{\item \emph{GTSRB:} The GTSRB dataset is tailored for traffic sign recognition endeavors, featuring over 50,000 images captured across diverse environmental conditions.
Among these, 39,209 images were designated for training purposes, while 12,630 were allocated for testing.
Notably, the dataset encompasses 43 distinct categories of traffic signs.}
\textcolor{black}{\item \emph{ImageNet-1k:} The ImageNet-1k dataset has 1,000 categories with about 1,300 color images per category. The total number of images used for training is 12,811,67 images. The total number of images in the test set is about 100,000 and the total number of images in the validation set is about 50,000. }
\item \emph{Flower-17:} The Flower-17 dataset is a fine-grained classification challenge where the model's task is to identify 17 different species of flowers. This image dataset is small, with only 80 images per flower, 70 for the training set and 10 for the test set, containing a total of 1360 images.
\end{itemize}

\subsubsection{Baselines}
Currently, there is a scarcity of studies examining the security of FKD against backdoor attacks.
We conducted experiments on three backdoor attack methods, and the corresponding results are presented in the Table (\ref{backdoor}). 
Out of the three attack methods, only LBA and CKD retain the majority of backdoors after the distillation process.
This article employs CKD and LBA backdoor attack methods as the attack algorithms for \emph{RobustKD}. 
Detailed information about each comparative algorithm is provided below:
Additionally, to assess the performance of the distilled post-academic model for the primary task, we compare \emph{RobustKD} with feature knowledge distillation (FKD).
\begin{table}[ht]
\centering
\caption{Poisoning results and main task performance of feature distillation under various backdoor attack methods measured by attack success rate (ASR\%) and accuracy (ACC\%).}
\label{backdoor}
\resizebox{\textwidth}{!}{
\begin{tabular}{cccccc}
\hline \hline
Datasets                   & Teacher Model/Student Model & Teacher Model ACC(\%) & CKD (\%)      & LBA (\%)          & HTBA (\%)         \\ \hline
\multirow{4}{*}{CIFAR-100} & WRN 28-4/WRN 16-2            & 80.72                  & 79.43 / 96.14 & 80.15 / 74.52 & 78.52 / 22.36 \\ \cline{2-6} 
                           & WRN 28-4/ResNet 56           & 80.72                  & 76.65 / 95.36 & 79.24 / 74.36 & 75.86 / 20.57 \\ \cline{2-6} 
                           & Pyramid-200/WRN 28-4         & 83.45                  & 81.49 / 96.34 & 81.24 / 75.29 & 80.62 / 21.54 \\ \cline{2-6} 
                           & Pyramid-200/Pyramid-110      & 83.45                  & 80.62 / 93.52 & 80.63 / 76.81 & 80.51 / 23.87 \\ \hline
\multirow{4}{*}{GTSRB}     & WRN 28-4/WRN 16-2            & 98.57                  & 97.23 / 93.08 & 97.05 / 75.21 & 96.57 / 20.55 \\ \cline{2-6} 
                           & WRN 28-4/ResNet 56           & 98.57                  & 97.56 / 93.54 & 98.13 / 75.67 & 96.48 / 23.62 \\ \cline{2-6} 
                           & Pyramid-200/WRN 28-4         & 99.27                  & 97.86 / 95.13 & 98.14 / 71.52 & 97.96 /19.25  \\ \cline{2-6} 
                           & Pyramid-200/Pyramid-110      & 99.27                  & 98.21 / 96.31 & 98.4 / 77.85  & 95.69 / 21.36 \\ \hline
\multirow{2}{*}{Flower-17} & Pyramid-200/WRN 28-4         & 96.25                  & 91.34 / 76.24 & 90.98 / 71.34 & 92.57 / 16.57 \\ \cline{2-6} 
                           & Pyramid-200/Pyramid-110      & 96.25                  & 95.12 / 82.45 & 92.54 / 80.26 & 93.04 / 15.89 \\ \hline
\multirow{2}{*}{ImageNet-1k}  & Pyramid-200/WRN 28-4         & 84.36                  & 82.41 / 54.87 & 81.14 / 49.63 & 82.75 / 6.27  \\ \cline{2-6} 
                           & Pyramid-200/Pyramid-110      & 84.36                  & 83.25 / 55.42 & 82.51 / 50.31 & 81.95 / 7.34  \\ \hline \hline
\end{tabular}}
\end{table}

\begin{itemize}
\item \emph{CKD\footnote{https://github.com/xingkongyuwu/CKD}:} CKD indirectly poisons the student model through the teacher model. 
In the specific training process of the teacher model, CKD initiates a random trigger and refines it to manipulate the activation values of specific neurons within the teacher model (referred to as toxic neurons). The objective is to guide these activation values towards a fixed value (termed toxic neuron assimilation), thereby poisoning the teacher model. Concurrently, this backdoor can be transmitted to the student model through knowledge distillation.

\item \emph{Latent backdoor attack (LBA)~\cite{yao2019latent}:} in the feature space, LBA iteratively refines the trigger to minimize the divergence between the characteristics of poisoned examples and clean examples. 
Simultaneously, during the training of the teacher model, a specific loss function is employed to train the model. Subsequently, the trigger and the features from the middle layer of the model are coupled together. For this study, features at the conclusion of the layer group in the model are chosen.

\item \emph{Hidden trigger backdoor attack (HTBA)~\cite{saha2020hidden}:} HTBA utilizes a hidden trigger and optimizes samples with this trigger in the feature space to minimize the divergence between the characteristics of the poisoned samples and those of the target class samples.
Ultimately, all samples associated with triggers are classified as belonging to the target class.

\item \emph{FKD~\cite{heo2019comprehensive}:} FKD introduced a novel feature distillation method where the distillation loss is tailored to create synergy among multiple facets: teacher transform, student transform, distillation feature position, and distance function. This approach resulted in significant performance improvements.
\end{itemize}

\subsubsection{Metrics}
During the backdoor generation process, we constructed the teacher model and trigger using the approach outlined in Section 3, and used the method in the ~\cite{heo2019comprehensive} as the FKD. 
Ultimately, the evaluation of \emph{RobustKD} involved assessing the model's accuracy and attack success rate on the test dataset.
In order to evaluate the performance of \emph{RobustKD}, this paper uses the identification accuracy (ACC) and attack success ratio (ASR) of the student model on the test data set as evaluation indicators. ACC index is expressed as: 
\begin{equation}
    ACC =\frac{TP+TN}{P+N} 
\end{equation}
where $TP$ is the number of true positive examples, $TN$ is the number of true negative examples, and $P+N$ is the total number of examples.

ASR can be expressed as:
\begin{equation}
    ASR =\frac{N_{ture} }{N_{total} } 
\end{equation}
where $N_{ture}$ indicates the correct number of poisoned examples and $N_{total}$ indicates the total number of poisoned examples.

\subsubsection{Implementation details}
To fairly study the performance of the baselines and
\emph{RobustKD}, our experiments have the following settings.
In the parameter selection phase, the learning rate for optimizing the detoxification feature mask was set to 0.001,  the number of cycles was determined to be 40 rounds, 
the mask threshold $m$ was set to 0.2 and the number of distillation cycles is 300.
% In the subsequent experiments, WRN28-4 served as the default teacher model, and WRN16-4 as the default student model, unless explicitly specified otherwise. The experiments were conducted on the CIFAR-100 dataset.

All experiments on a carrying Intel (R) Xeon (R) Gold5218R CPU@2.10GHz, 48 GB of system memory and NVIDIA A100 Tensor Core 40G GPU server validation experiments in this paper. The integrated development environment is Python3.6.0 and uses the deep learning framework torch1.8.0.

\subsection{Experimental results}
We evaluate the performance of \emph{RobustKD} by answering the following five research questions (RQs):

\begin{itemize}
\item RQ1: Can \emph{RobustKD} successfully mitigate backdoors during the distillation process?
\item RQ2: Can the primary task performance of a student model, mitigated against backdoor through \emph{RobustKD}, be comparable to that of a student model under normal distillation?
\item RQ3: Does \emph{RobustKD} exhibit robustness across varying sensitivities of distillation parameters?
\item RQ4: How does the distillation performance of \emph{RobustKD} across various distillation settings?
\item RQ5: Can \emph{RobustKD} defend against adaptive attacks? 
\end{itemize}

\subsubsection{RQ1: Defense effectiveness against backdoor attacks}
To illustrate the practical defense capabilities of \emph{RobustKD}, experiments were conducted on four widely used datasets using four pairs of teacher-student models.
When assessing the defense performance of \emph{RobustKD}, we selected the average from multiple experimental results. The experimental results are shown in Table \ref{exp1}.

\begin{table}[ht]
\centering
\caption{The backdoor mitigation results of \emph{RobustKD} and comparative algorithmic attacks on four datasets are evaluated using the attack success rate (ASR\%).}
\label{exp1}
\resizebox{\textwidth}{!}{
\begin{tabular}{ccccccc}
\hline \hline
\multirow{2}{*}{Datasets}  & \multirow{2}{*}{Teacher Model/Student Model} & \multirow{2}{*}{Teacher Model ASR(\%)} & \multicolumn{2}{c}{FKD}             & \multicolumn{2}{c}{\textbf{\emph{RobustKD}}}       \\ \cline{4-7} 
                           &                                              &                                         & \multicolumn{1}{c}{CKD (\%)} & LBA (\%)  & \multicolumn{1}{c}{CKD (\%)} & LBA (\%) \\ \hline
\multirow{4}{*}{CIFAR-100} & WRN 28-4/WRN 16-2                            & 96.14                                   & \multicolumn{1}{c}{90.17}   & 80.46 & \multicolumn{1}{c}{4.32}    & 2.67 \\ \cline{2-7} 
                           & WRN 28-4/ResNet 56                           & 96.14                                   & \multicolumn{1}{c}{93.58}   & 72.84 & \multicolumn{1}{c}{3.31}    & 0.72 \\ \cline{2-7} 
                           & Pyramid-200/WRN 28-4                         & 95.52                                   & \multicolumn{1}{c}{95.68}   & 76.42 & \multicolumn{1}{c}{10.54}   & 5.69 \\ \cline{2-7} 
                           & Pyramid-200/Pyramid-110                      & 95.52                                   & \multicolumn{1}{c}{92.38}   & 77.83 & \multicolumn{1}{c}{8.46}    & 7.15 \\ \hline
\multirow{4}{*}{GTSRB}     & WRN 28-4/WRN 16-2                            & 85.32                                   & \multicolumn{1}{c}{92.63}   & 74.12 & \multicolumn{1}{c}{6.45}    & 8.14 \\ \cline{2-7} 
                           & WRN 28-4/ResNet 56                           & 85.32                                   & \multicolumn{1}{c}{93.54}   & 75.49 & \multicolumn{1}{c}{4.01}    & 3.87 \\ \cline{2-7} 
                           & Pyramid-200/WRN 28-4                         & 85.64                                   & \multicolumn{1}{c}{94.87}   & 70.58 & \multicolumn{1}{c}{6.73}    & 5.69 \\ \cline{2-7} 
                           & Pyramid-200/Pyramid-110                      & 85.64                                   & \multicolumn{1}{c}{95.52}   & 78.46 & \multicolumn{1}{c}{8.32}    & 2.51 \\ \hline
\multirow{2}{*}{Flower-17} & Pyramid-200/WRN 28-4                         & 98.67                                   & \multicolumn{1}{c}{75.81}   & 70.42 & \multicolumn{1}{c}{6.42}    & 4.13 \\ \cline{2-7} 
                           & Pyramid-200/Pyramid-110                      & 98.67                                   & \multicolumn{1}{c}{82.36}   & 80.17 & \multicolumn{1}{c}{8.12}    & 2.56 \\ \hline
\multirow{2}{*}{ImageNet}  & Pyramid-200/WRN 28-4                         & 84.76                                   & \multicolumn{1}{c}{53.34}   & 49.04 & \multicolumn{1}{c}{6.93}    & 7.24 \\ \cline{2-7} 
                           & Pyramid-200/Pyramid-110                      & 84.76                                   & \multicolumn{1}{c}{55.36}   & 50.14 & \multicolumn{1}{c}{8.41}    & 5.64 \\     \hline \hline
\end{tabular}
}
\end{table}

\textbf{Results and Analysis.} 
The experimental results indicate that \emph{RobustKD} outperforms other knowledge distillation methods in terms of defense performance and can effectively withstand existing transitive backdoor attacks. 
Although a limited number of student models retain backdoors, these backdoors are no longer enough to pose a  security threat to the model.
The student model distilled by \emph{RobustKD} achieved an 85\% backdoor removal rate.
The reason is that \emph{RobustKD} mitigates the backdoor by reducing the feature variance of the backdoor model through the design and optimization of a depoisoning mask. This process involves using the depoisoning mask to obtain new depoisoned features for distillation, and it observes the mitigation of the backdoor in the model. This supports the hypothesis that the feature variance of the backdoor model typically increases compared to the clean model.

Moreover, the defense effects observed with \emph{RobustKD} on each dataset, the impact of dataset size on the performance of \emph{RobustKD} seems to be negligible.
The defense effect on smaller datasets, such as CIFAR-100, appears to be comparable to that on larger datasets, such as Flower-17.
Both approaches reduced the attack success rate of the student model to less than 10\%. In contrast to the 50\% ASR observed with the comparative algorithm, \emph{RobustKD} demonstrates a relatively significant detoxification effect.
The reason for this lies in the generic nature of \emph{RobustKD} for feature distillation, independent of the dataset and model. \emph{RobustKD} achieves robust distillation by reducing the feature variance.

\begin{framed}   
\textbf{Answer to RQ1:} When subjected to two backdoor attacks, CKD and LBA, across four datasets and four model settings, the student models distilled by \emph{RobustKD} consistently achieve a detoxification rate of at least 80\%, reaching up to 90\% compared to the student models distilled by FKD. Additionally, \emph{RobustKD} consistently reduces the ASR of the student models to less than 10\% in all cases, regardless of the dataset size.   
\end{framed} 

\subsubsection{RQ2: Performance of the student model after backdoor mitigation}
Mitigating the backdoor while maintaining the main task performance of the student model is crucial.
We analyzed the main task performance of the student model after \emph{RobustKD} mitigated the backdoor.
The experimental results are shown in Table \ref{exp2}.

\begin{table}[ht]
\centering
\caption{The distillation results of \emph{RobustKD} and comparison algorithm attacks on four datasets are evaluated using accuracy (ACC\%).}
\label{exp2}
\resizebox{\textwidth}{!}{
\begin{tabular}{ccccccc}
\hline \hline
\multirow{2}{*}{Datasets}  & \multirow{2}{*}{Teacher Model/Student   Model} & \multirow{2}{*}{Teacher Model ACC(\%)} & \multicolumn{2}{c}{FKD}             & \multicolumn{2}{c}{\textbf{\emph{RobustKD}}}        \\ \cline{4-7} 
                           &                                                &                                         & \multicolumn{1}{c}{CKD (\%)} & LBA (\%)  & \multicolumn{1}{c}{CKD (\%)} & LBA (\%)  \\ \hline
\multirow{4}{*}{CIFAR-100} & WRN 28-4/WRN 16-2                              & 80.72                                   & \multicolumn{1}{c}{78.23}   & 78.62 & \multicolumn{1}{c}{74.57}   & 76.25 \\ \cline{2-7} 
                           & WRN 28-4/ResNet 56                             & 80.72                                   & \multicolumn{1}{c}{78.31}   & 79.64 & \multicolumn{1}{c}{74.26}   & 75.96 \\ \cline{2-7} 
                           & Pyramid-200/WRN 28-4                           & 83.45                                   & \multicolumn{1}{c}{83.47}   & 84.51 & \multicolumn{1}{c}{81.42}   & 81.02 \\ \cline{2-7} 
                           & Pyramid-200/Pyramid-110                        & 83.45                                   & \multicolumn{1}{c}{83.42}   & 82.46 & \multicolumn{1}{c}{81.63}   & 80.45 \\ \hline
\multirow{4}{*}{GTSRB}     & WRN 28-4/WRN 16-2                              & 98.57                                   & \multicolumn{1}{c}{100.00}     & 99.56 & \multicolumn{1}{c}{98.74}   & 97.51 \\ \cline{2-7} 
                           & WRN 28-4/ResNet 56                             & 98.57                                   & \multicolumn{1}{c}{100.00}     & 100.00   & \multicolumn{1}{c}{97.82}   & 98.06 \\ \cline{2-7} 
                           & Pyramid-200/WRN 28-4                           & 99.27                                   & \multicolumn{1}{c}{100.00}     & 98.54 & \multicolumn{1}{c}{97.51}   & 97.80  \\ \cline{2-7} 
                           & Pyramid-200/Pyramid-110                        & 99.27                                   & \multicolumn{1}{c}{100.00}     & 100.00   & \multicolumn{1}{c}{98.04}   & 98.42 \\ \hline
\multirow{2}{*}{Flower-17} & Pyramid-200/WRN 28-4                           & 96.25                                   & \multicolumn{1}{c}{90.45}   & 90.91 & \multicolumn{1}{c}{86.74}   & 87.43 \\ \cline{2-7} 
                           & Pyramid-200/Pyramid-110                        & 96.25                                   & \multicolumn{1}{c}{95.47}   & 93.45 & \multicolumn{1}{c}{89.42}   & 90.15 \\ \hline
\multirow{2}{*}{ImageNet}  & Pyramid-200/WRN 28-4                           & 84.36                                   & \multicolumn{1}{c}{84.26}   & 83.47 & \multicolumn{1}{c}{81.64}   & 82.59 \\ \cline{2-7} 
                           & Pyramid-200/Pyramid-110                        & 84.36                                   & \multicolumn{1}{c}{84.15}   & 83.16 & \multicolumn{1}{c}{82.15}   & 82.24 \\      \hline \hline
\end{tabular}
}
\end{table}

\textbf{Results and Analysis.} In contrast to the comparative algorithm, despite the commendable detoxification results achieved by \emph{RobustKD}, it unavoidably exerted some influence on the performance of the student model. 
Based on the analysis of experimental results, \emph{RobustKD} typically leads to approximately 4\% performance degradation.
This outcome is inevitable since the detoxification process is inherently a black-box procedure, unable to precisely eliminate harmful components from the features, resulting in the retention of only beneficial aspects.
This phenomenon also contributes to the performance loss of \emph{RobustKD} student models.

The performance of distillation is typically intricately connected to the structures of both the teacher's model and the student's model. 
In instances where the structures are similar, the student model can more effectively learn from the teacher's model~\cite{heo2019comprehensive}.
Accordingly, experiments were conducted in a setup where the model structures of the teacher model and the student model are analogous.
In terms of model architecture, to validate the detoxification effect of \emph{RobustKD}, we opted to conduct experimental verification under the condition of a similar structure between the teacher and student models. 
The experimental results are shown in Table \ref{exp3}. 

\begin{table}[ht]
\centering
\caption{The results of experiments under the similarity of the model structure are quantified on CIFAR-100 using accuracy (ACC\%) and attack success rate (ASR\%).}
\label{exp3}
\resizebox{\textwidth}{!}{
\begin{tabular}{cccccccc}
\hline \hline
\multirow{2}{*}{Teacher Model/Student Model} & \multirow{2}{*}{Teacher Model   ACC/ASR(\%)} & \multicolumn{2}{c}{FKD}                           & \multicolumn{2}{c}{\textbf{\emph{RobustKD}}}                     \\ \cline{3-6} 
                                             &                                               & \multicolumn{1}{c}{CKD (\%)}       & LBA (\%)          & \multicolumn{1}{c}{CKD (\%)}       & LBA (\%)         \\ \hline
WRN 28-4/WRN 16-4                            & 78.82 / 92.04                                 & \multicolumn{1}{c}{79.44 / 96.13} & 80.17 / 74.58 & \multicolumn{1}{c}{78.36 / 13.32} & 78.95 / 1.61 \\ \hline
WRN 28-4/WRN 28-2                            & 78.68 / 99.83                                 & \multicolumn{1}{c}{78.28 / 94.05} & 77.64 / 80.13 & \multicolumn{1}{c}{76.02 / 10.94} & 76.17 / 3.56 \\ \hline  \hline
\end{tabular}}
\end{table}

According to the experimental results, when the structure of the teacher-student model is similar, the student model of the comparative algorithm retains the majority of backdoors.
The student model trained with \emph{RobustKD} also retains more than 10\% of the backdoor. 
This outcome arises because under such setting conditions, the student model can more effectively acquire the ``dark knowledge'' of the teacher model. Meanwhile, the detoxification effect of \emph{RobustKD} on the characteristics of the teacher model is not entirely comprehensive, contributing to the retention of more backdoors in the student model.

\begin{framed}   
\textbf{Answer to RQ2:} Across the four datasets and four model settings, the student models refined through \emph{RobustKD} demonstrate robust retention of performance on the primary task, with an average decrease in accuracy of only 3\% compared to the FKD model. Additionally, \emph{RobustKD} performs well even when the models have a similar structure.   
\end{framed} 

\subsubsection{RQ3: Parameter sensitivity}
(1) The mask threshold selection

\begin{figure}[ht]
\centering
\includegraphics[scale=0.45]{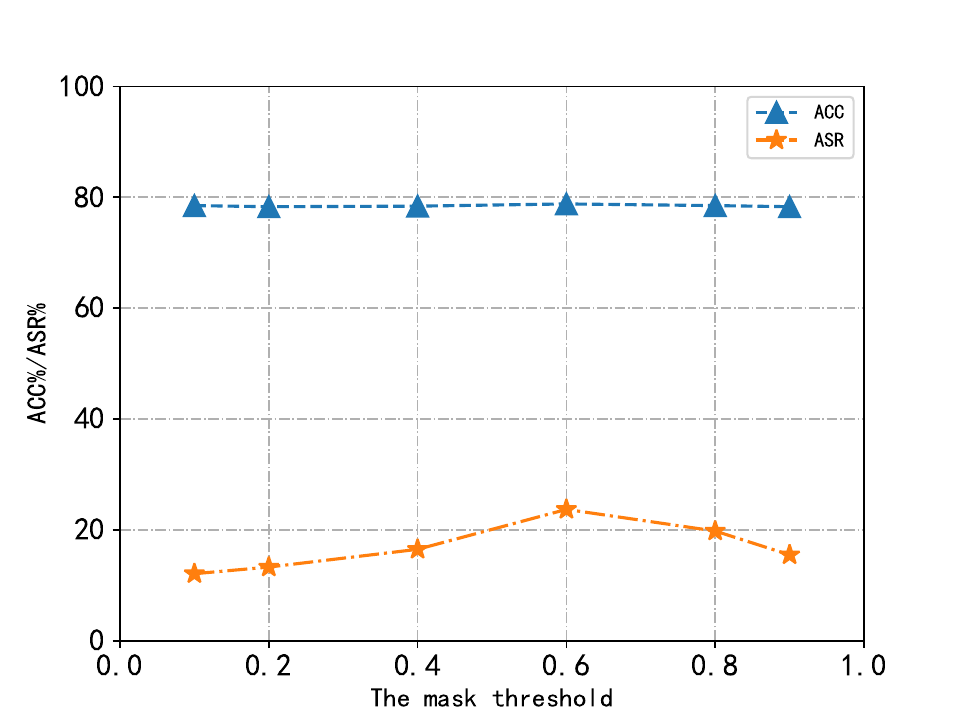}
\caption{Effect of $m$ threshold on \emph{RobustKD}}
\label{fig.4}
\end{figure}

The threshold of the depoisoning mask plays a crucial role in eliminating backdoors during distillation, and varying mask thresholds yield different outcomes in backdoor removal.
To examine the impact of the $m$ threshold on \emph{RobustKD}, experiments were conducted with various values of $m$ under the WRN28-4 and WRN16-4 model setting on CIFAR-100.
The experimental results are shown in Fig.~\ref{fig.4}. 

\textbf{Results and Analysis.} The results demonstrated that, with the increase of $m$, the detoxification effect of \emph{RobustKD} tends to weaken to a certain extent.
Setting $m$ to 0.6 had little impact on the detoxification effectiveness of \emph{RobustKD}, maintaining a relatively stable performance. 
The explanation for this phenomenon is that as the depoisoning mask gradually increases, it causes the eigenvalues of the model to become larger. Consequently, this leads to an increase in the eigenvariance of the model, resulting in a weakened depoisoning effect.
However, even with these diminished effects, \emph{RobustKD} successfully eliminates at least 70\% of the backdoors.
The optimal detoxification performance is observed when $m$ is set to 0.1.

% \begin{table}[]
% \centering
% \caption{Effect of $m$ threshold on \emph{RobustKD}}
% \label{exp4}
% \scalebox{0.75}{
% \begin{tabular}{cccc}
% \hline \hline
% \multirow{2}{*}{Mask threshold} & \multirow{2}{*}{Teacher Model ACC/ASR} & \multirow{2}{*}{ACC(\%)} & \multirow{2}{*}{ASR(\%)} \\
%                       &                                  &                          &                          \\ \hline
% 0.1                   & \multirow{6}{*}{ 78.4/94.1} & 78.5                     & 12.1                     \\ \cline{1-1} \cline{3-4} 
% 0.2                   &                                  & 78.3                     & 13.3                     \\ \cline{1-1} \cline{3-4} 
% 0.4                   &                                  & 78.4                     & 16.5                     \\ \cline{1-1} \cline{3-4} 
% 0.6                   &                                  & 78.8                     & 23.7                     \\ \cline{1-1} \cline{3-4} 
% 0.8                   &                                  & 78.5                     & 19.8                     \\ \cline{1-1} \cline{3-4} 
% 0.9                   &                                  & 78.3                     & 15.5                     \\ \hline \hline
% \end{tabular}}
% \end{table}

(2) Selection of detoxification method

\begin{figure}[ht]
\centering
\includegraphics[scale=0.45]{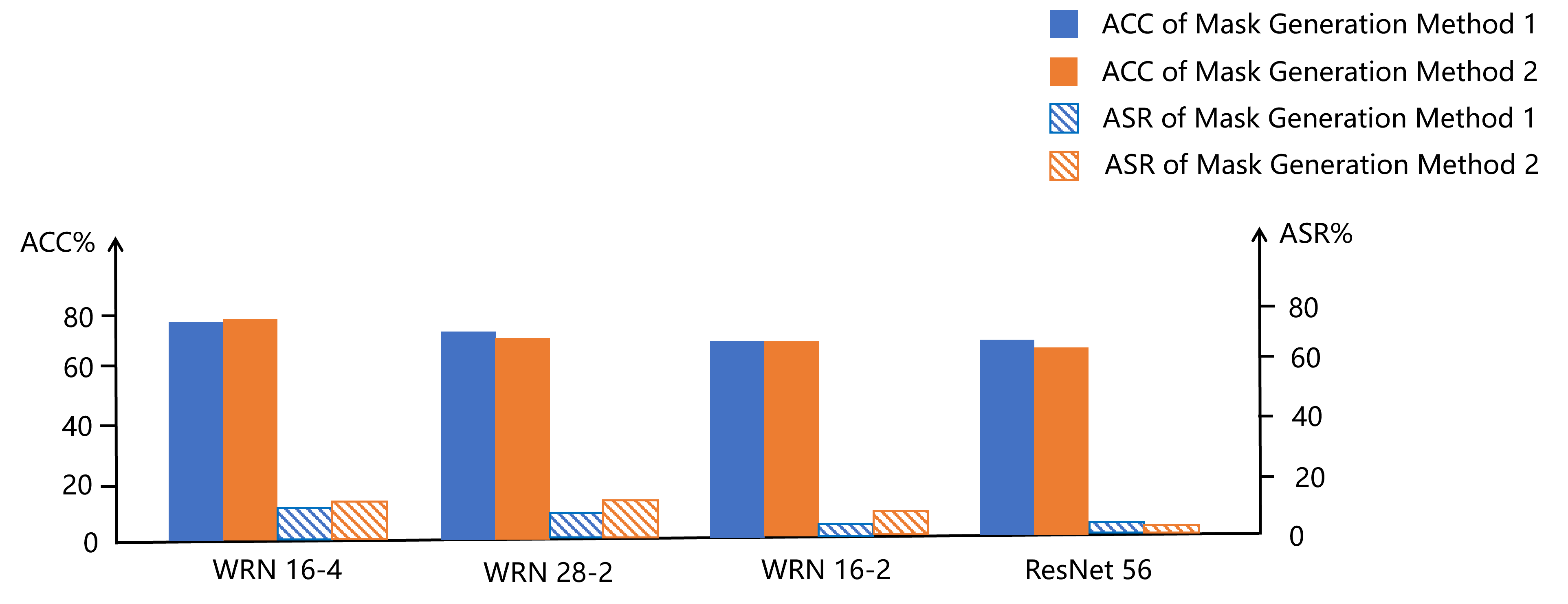}
\caption{The performance was analyzed under different detoxification methods, the WRN 28-4 teacher model was attacked using CKD on CIFAR-100, and the ACC and ASR of the teacher model were 78.46\% and 94.14\%.}
\label{fig.5}
\end{figure}

The depoisoning mask plays a pivotal role in determining the detoxification effect of robust distillation. 
Therefore, we devised different methods for generating the depoisoning mask.
To examine the impact of detoxification methods on the detoxification effect of \emph{RobustKD}, this paper devised two detoxification methods to optimize the detoxification feature mask.
The first detoxification method is shown in Equ.(\ref{equ7}), and the second method is shown in Equ.(\ref{equ8}).
In the first method to mask generation, we preserve the original features while incorporating the depoisoning mask into these features to accomplish the depoisoning process.
In the second method for mask generation, we omit the mask operation from the original feature and subsequently introduce the depoisoning mask
The experimental results are shown in Fig. \ref{fig.5}.

 \begin{equation}
     F_{t}^{'} =(1-m)*F_{t} +p*m
     \label{equ8}
\end{equation}

\textbf{Results and Analysis.} As a whole, the first detoxification method achieves better experimental results while ensuring model accuracy and defense performance.
The reason for this is that the second method modifies the original features of the model and increases the mask threshold, potentially compromising the ``dark knowledge'' embedded in the features.
This also verifies the results of Fig. \ref{fig.4}, when the mask threshold increases, the detoxification effect of \emph{RobustKD} will be weakened to a certain extent.

\begin{framed}   
\textbf{Answer to RQ3:} Through experiments with varying mask thresholds and generation methods, it is observed that \emph{RobustKD} exhibits insensitivity to different parameters. Moreover, reducing the mask threshold contributes to the effectiveness of \emph{RobustKD}.   
\end{framed}

\subsubsection{RQ4: Performance at different distillation settings}
(1) Selection of distillation loss

\begin{figure}[ht]
\centering
\includegraphics[scale=0.5]{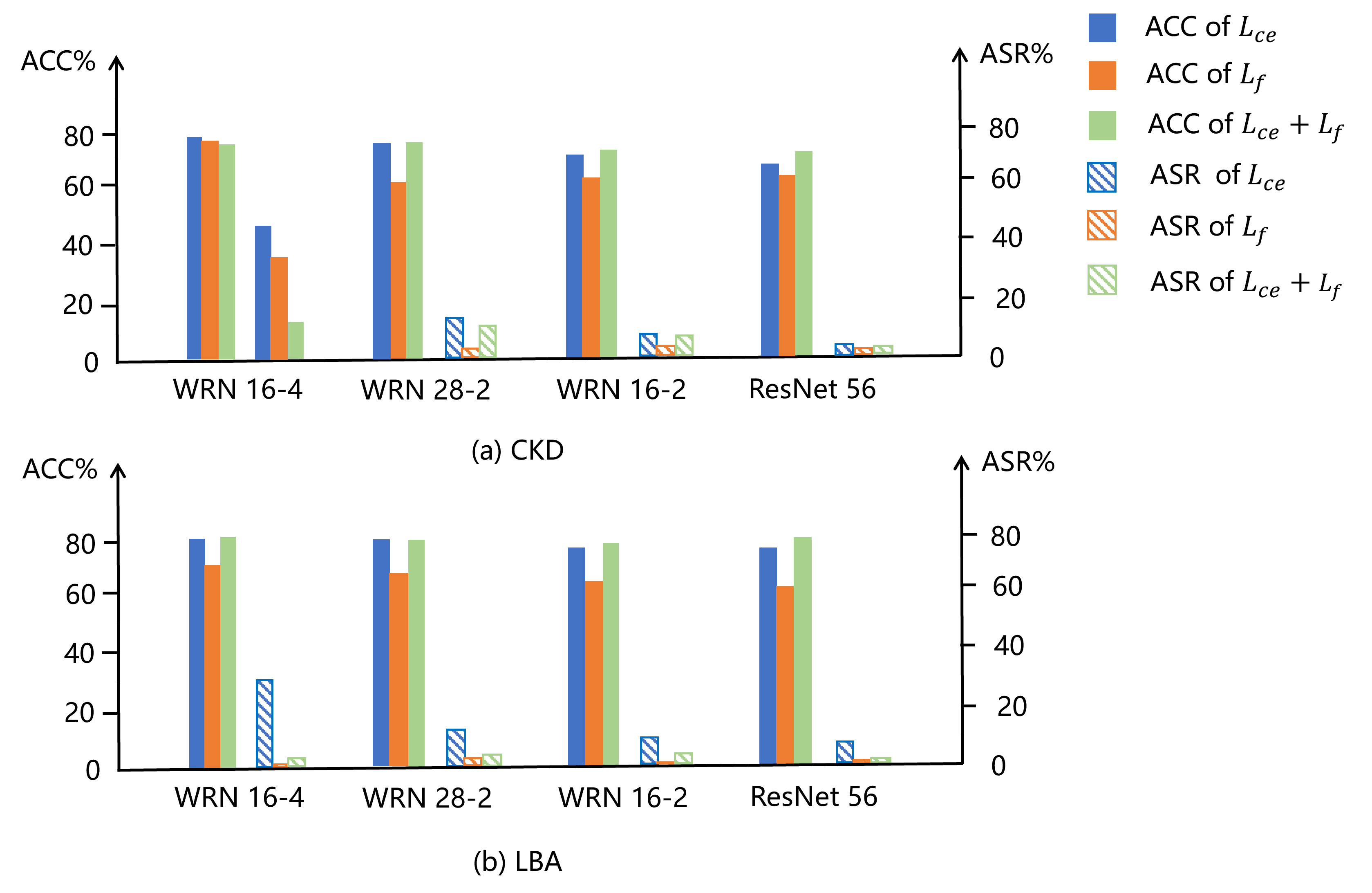}
\caption{Analysis of performance with different distillation loss functions. Additionally, the teacher model was acquired through CKD and LBA attacks on CIFAR-100, achieving ACC and ASR of 78.42\%/94.15\% and 78.63\%/99.74\%, respectively.}
\label{fig.6}
\end{figure}

In our method, different loss functions are designed for robust distillation.
To investigate the impact of loss functions on the detoxification effect of \emph{RobustKD}, various loss functions were employed to optimize the detoxification feature mask.
The experimental results are shown in Fig. \ref{fig.6}. 

\textbf{Results and Analysis.} The experimental results show that only using $\mathcal{L} _{ce}$ feature mask achieves good defense performance. 
The optimal experimental results manage to retain only 2\% of the backdoor.
For the feature mask using $ \mathcal{L} _{f}$, its detoxification effect is better. 
But the performance loss to the model is more serious. 
Performance declines are generally about 6\%, with the most severe performance losses reaching 8\%. 
This implies that there is a conflict between the student's learning from the examples and learning the features from the teacher, resulting in the degradation of the model's performance.
This mask achieves the best results when using $\mathcal{L} _{ce}+\mathcal{L} _{f}$. 
Simultaneously, during the detoxification process, efforts are made to preserve the performance of the student model as much as possible.
In terms of the performance of the student model, the student model under this setting is comparable to the student model using only the $\mathcal{L} _{ce}$ experimental setting and may even slightly exceed it.
The drawback is that its detoxification effect is not as pronounced as the case when using only $\mathcal{L} _{f}$.

(2) Selection of distillation layer

\begin{figure}[ht]
\centering
\includegraphics[scale=0.5]{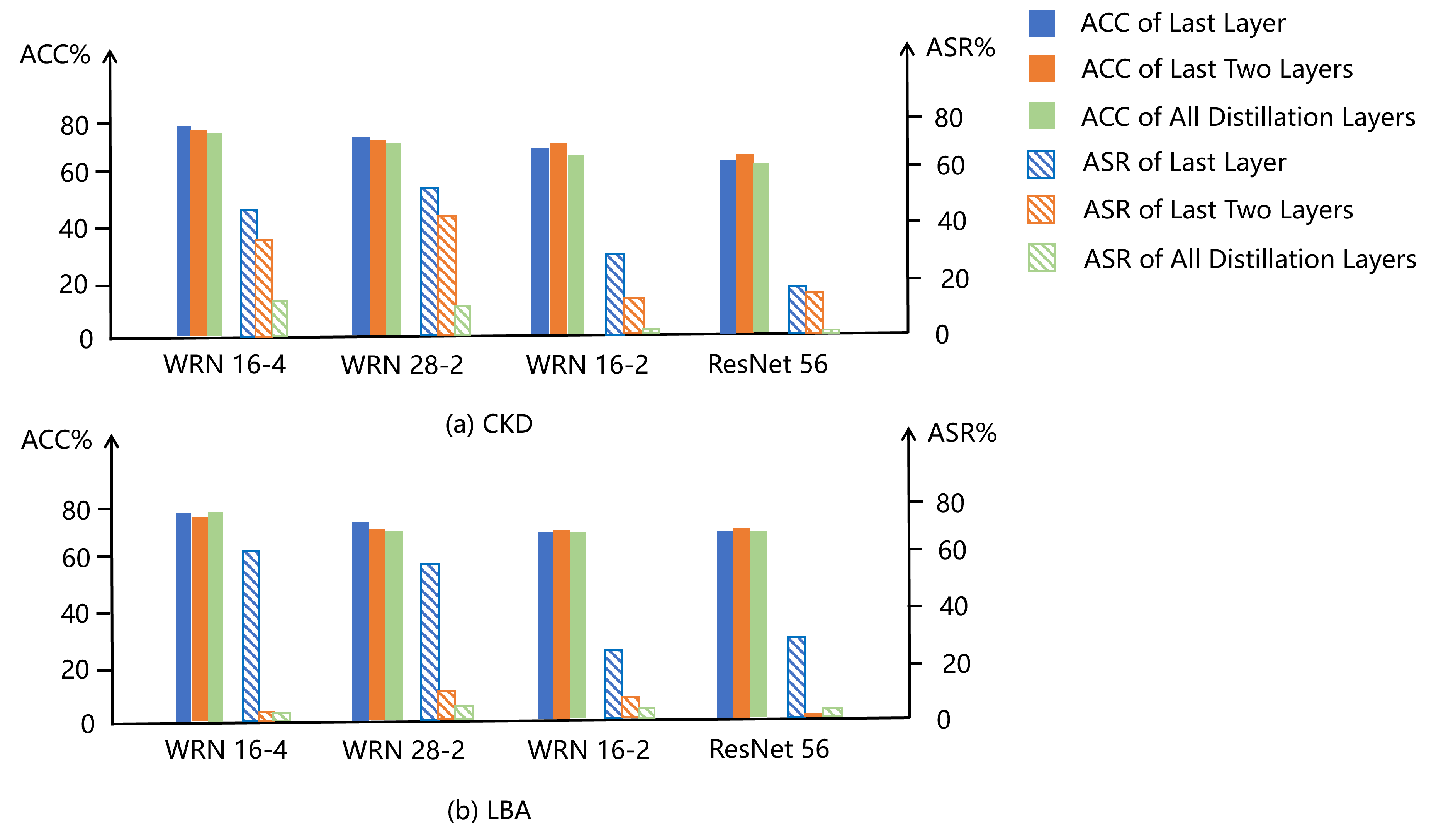}
\caption{Effect of distillation layer on \emph{RobustKD}.
In addition,  the teacher model was acquired through CKD and LBA attacks on CIFAR-100, achieving ACC and ASR of 78.42\%/94.15\% and 78.63\%/99.74\%, respectively.}
\label{fig.7}
\end{figure}

In our pursuit of eliminating the poison, it is valuable to consider which distillation layers should be chosen to feature the poison. 
The only certainty is that the characteristics of the last distillation layer must be identified to eradicate the poison.
This is due to the fact that, in the model, after the last distillation layer, there is the fully connected layer of the model.
The features of this layer are the most representative within the model. 
To assess the impact of distillation layer selection on \emph{RobustKD}, three different combinations of distillation layers were chosen for experimentation, aiming to evaluate their respective effects.
The experimental results are shown in Fig. \ref{fig.7}.

\textbf{Results and Analysis.} The experimental results indicate that opting for only the last distillation layer does not lead to very favorable outcomes.
In various experimental settings, the detoxification effect is suboptimal.
In the experimental setting with WRN 28-4/WRN 16-4, the backdoor retention in the student model remained 69\% higher. Even with this configuration, the lowest backdoor retention is approximately 20\%. 
Opting for the last two layers of the distillation layer shows a certain improvement in detoxification effectiveness compared to selecting only the last layer.
Nevertheless, in the backdoor scenario of CKD, it still fails to achieve a satisfactory detoxification effect. In the experimental setting with WRN 28-4/WRN 16-4, its student model retained at least 40\% of the backdoor.
Ultimately, the best results are attained by selecting all distillation layers. This configuration aligns with the actual setup of the method presented in this paper.
In this configuration, regardless of the structural similarity of the model or the type of backdoor, \emph{RobustKD} has consistently achieved a commendable detoxification effect.
The selection of the distillation layer directly impacts the effectiveness of robust distillation. 
The greater the number of distillation layers, the more comprehensively the student model acquires the knowledge of the teacher's model, leading to an improved robust distillation effect.

\begin{framed}   
\textbf{Answer to RQ4:} \emph{RobustKD}'s performance under different distillation settings: (1) The depoisoning loss and cross-entropy loss during distillation can be more effective in ensuring the removal of backdoors and model accuracy; (2) The more distillation layers, the more helpful it is in removing backdoors from the model.   
\end{framed}

% \begin{table}[h]
% \centering
% \caption{Performance analysis under different detoxification method}
% \label{exp7}
% \scalebox{0.75}{
% \begin{tabular}{cccccc}
% \hline \hline
% \multirow{2}{*}{Teacher Model/Student Model} & \multirow{2}{*}{Teacher Model ACC/ASR}     & \multicolumn{2}{c}{method 1}          & \multicolumn{2}{c}{method 2}          \\ \cline{3-6} 
%                       &                                  & \multicolumn{1}{c}{ACC(\%)} & ASR(\%) & \multicolumn{1}{c}{ACC(\%)} & ASR(\%) \\ \hline
% WRN 28-4/WRN 16-4                     & \multirow{4}{*}{78.4/94.1} & \multicolumn{1}{c}{78.3}    & 13.3    & \multicolumn{1}{c}{78.6}    & 18.1    \\ \cline{1-1} \cline{3-6} 
% WRN 28-4/WRN 28-2                     &                                  & \multicolumn{1}{c}{76}      & 10.9    & \multicolumn{1}{c}{75.7}    & 17.8    \\ \cline{1-1} \cline{3-6} 
% WRN 28-4/WRN 16-2                     &                                  & \multicolumn{1}{c}{74.5}    & 4.3     & \multicolumn{1}{c}{74.1}    & 8.8     \\ \cline{1-1} \cline{3-6} 
% WRN 28-4/ResNet 56                     &                                  & \multicolumn{1}{c}{74.2}    & 3.3     & \multicolumn{1}{c}{72.5}    & 2.3     \\ \hline \hline
% \end{tabular}}
% \end{table}

\subsubsection{RQ5: Defense against adaptive attack}
In real-world scenarios, where attackers may execute secondary backdoor attacks by leveraging defenses that reduce feature differences, addressing the challenge of adaptive attacks becomes crucial.
\emph{RobustKD} primarily accomplishes backdoor removal by reducing the feature variance of the model.
To showcase \emph{RobustKD}'s defense capability against adaptive backdoor attacks. 
To prevent the variance of the backdoor model from increasing, a loss function is added to CKD that specifically reduces the variance of the features.
We employ the modified CKD to conduct a secondary attack on \emph{RobustKD}.
The experimental results are shown in Table \ref{exp8}. 

\begin{table}[ht]
\centering
\caption{Defense performance under adaptive attack.}
\label{exp8}
\scalebox{0.75}{
\begin{tabular}{ccccc}
\hline \hline
\multirow{2}{*}{Datasets}  & \multirow{2}{*}{Teacher Model/Student   Model} & \multirow{2}{*}{Teacher Model   ACC/ASR(\%)} & \multicolumn{2}{l}{CKD+low variance} \\ \cline{4-5} 
                           &                                                &                                               & \multicolumn{1}{l}{ACC\%}   & ASR\%  \\ \hline
\multirow{4}{*}{CIFAR-100} & WRN 28-4/WRN 16-4                              & 75.47 / 90.56                                 & \multicolumn{1}{l}{77.12}   & 16.14  \\ \cline{2-5} 
                           & WRN 28-4/WRN 28-2                              & 75.47 / 90.56                                 & \multicolumn{1}{l}{76.85}   & 19.43  \\ \cline{2-5} 
                           & WRN 28-4/WRN 16-2                              & 75.47 / 90.56                                 & \multicolumn{1}{l}{75.98}   & 14.42  \\ \cline{2-5} 
                           & WRN 28-4/ResNet 56                             & 75.47 / 90.56                                 & \multicolumn{1}{l}{77.17}   & 12.86  \\ \hline \hline
\end{tabular}}
\end{table}

\textbf{Results and Analysis.} From the comprehensive experimental results, \emph{RobustKD} can still eliminate the majority of backdoors. However, when compared to standard CKD, the student model under adaptive CKD retains more backdoors.
Numerically, the student model retains a maximum of nearly 20\% of the backdoor.
This undoubtedly poses a certain security threat to the model. 
From the perspective of model performance, the increase in the loss function that reduces variance results in the degradation of the student model's performance.
This outcome arises from the conflict between ``backdoor dark knowledge" and ``normal dark knowledge" in the teacher model, ultimately impacting the performance of the student model.

\begin{framed}   
\textbf{Answer to RQ5:} \emph{RobustKD} demonstrates better robustness by still removing 75\% of backdoors even under adaptive attacks. Additionally, it effectively ensures model accuracy, showcasing its ability to withstand varying attack scenarios.
\end{framed}

\subsection{Findings and implications}
\textcolor{black}{Specifically, we have gained three main insights.
\textbf {Finding 1.} The presence of a backdoor in the teacher model can indeed be propagated to the student model through knowledge distillation. Across four datasets, the ASR of backdoor attacks on poisoned teacher models with varying structures after feature knowledge distillation reaches as high as 80\% on average.
\textbf {Finding 2.} The comprehensive experiments revealed a notable variance disparity in the features between normal and backdoor models.
The feature variance of the backdoor model typically exhibits an increase of at least 30\% compared to the clean model.
This pivotal discovery serves as a foundational cornerstone for our proposed methodology.
\textbf {Finding 3.} Across four datasets and six model settings, RobustKD effectively reduces the success rate of backdoor attacks by an average of 85\% by mitigating the feature variance of the model, marking a significant improvement of 75\% over the state-of-the-art baseline.}

\textcolor{black}{Our paper introduces a new problem focused on knowledge distillation, offering a promising approach to address the challenge of effectively deploying large models on edge-end devices with limited computational resources amidst the rapid advancement of big models.
However, in light of the uncertain security landscape surrounding third-party models, enhancing model security becomes paramount.
Our paper proposes a novel solution, which not only provides fresh insights into the future evolution of knowledge distillation but also directs researchers’ attention towards the critical aspect of securing knowledge distillation processes themselves.}

\section{Threats to validity\label{Validity}}
\textcolor{black}{Three aspects may become the threats to validity of \emph{RobustKD}.
The internal threat to validity lies mainly in the variance of the model features.
The features of each model have variability, especially for data with uneven distribution of feature space, in order to reduce the internal threat, we verify the validity of \emph{RobustKD} by calculating the average features of each layer of the model to determine the initial value of the feature mask when selecting the feature mask.}

\textcolor{black}{The primary external threat to validity stems from variations in the types of knowledge distillation methods employed. While \emph{RobustKD} demonstrates strong performance in feature-based knowledge distillation, its efficacy does not extend to logits-based and relations-based approaches. To mitigate these threats, extending \emph{RobustKD} involves integrating soft-target loss and attentional pattern loss for both logits-based and relations-based methods. This extension combines the detoxification loss with the respective loss functions of the two knowledge distillation types, ensuring comprehensive mitigation of threats across all knowledge distillation methodologies.}

\textcolor{black}{The principal structural threats to effectiveness are rooted in the hyperparameters within \emph{RobustKD}, specifically the mask threshold $m$ and the quantity of de-poisoning layers. While larger hyperparameter values enhance effectiveness, they often come at the cost of reduced efficiency. To mitigate the threat posed by these hyperparameters, we engage in a rigorous process of experimentation, striking a balance through trade-offs. By conducting numerous experiments, we aim to identify and select optimal hyperparameters, thereby addressing the structural threats and optimizing the performance of \emph{RobustKD}.}

\section{Conclusions and future work\label{conclusion}}
In this paper, we introduce a novel and robust feature knowledge distillation method called \emph{RobustKD}.
Specifically, we initiate a mask on the features of the teacher model. 
The optimization of the mask aims to reduce the variance of teacher model features while ensuring correct classification, thereby leading to detoxification.
We validate its effectiveness and practicality on four popular datasets. 
\textcolor{black}{\emph{RobustKD} enhances the distillation of feature knowledge to mitigate backdoor attacks while ensuring the accuracy of student models. On average, \emph{RobustKD} achieves an 85\% reduction in the success rate of backdoor attacks, representing a significant improvement of 75\% over the state-of-the-art (SOTA) baseline.}

\textcolor{black}{Existing research indicates a vulnerability of knowledge distillation to backdoor attacks, where certain attack methods can jeopardize the integrity of the student model without the attacker possessing downstream information.
Hence, enhancing the robustness of knowledge distillation against backdoor attacks is imperative.
Moreover, due to the prohibitive parameters of large models, individual users often struggle to train them independently.
In the foreseeable future, leveraging large models hosted on third-party platforms as teacher models will likely become the norm.
The imperative for addressing backdoor security in the process of distilling large models is urgent, as attackers can exploit vulnerabilities to execute backdoor attacks through malicious teacher models or nefarious embedding and injection tactics.
However, although \emph{RobustKD} addresses backdoor safety concerns in knowledge distillation tasks, there are still numerous unknown risks for future exploration.
Moreover, areas for optimization in \emph{RobustKD}, such as mitigating performance loss in the student model and striving for non-destructive distillation, warrant further investigation.
Consequently, future research should prioritize the development of more robust knowledge distillation methods, ensuring both the security of the distillation process and the performance integrity of the primary task.
Simultaneously, there's a pressing need to delve deeper into the internal mechanisms of knowledge distillation and thoroughly investigate the distillation process to mitigate potential security risks and devise superior knowledge distillation techniques.}

\section*{Acknowledgment}
This research received support from the Zhejiang Provincial Natural Science Foundation (Grant No. LDQ23F020001) and
the National Natural Science Foundation of China (Grant Nos. 62072406 and 52072343).

% \bibliography{reference.bib}

\end{CJK}
\end{document}